# Behavioral Anomaly Detection in Distributed Systems via Federated Contrastive Learning

Renzi Meng
Northeastern University
Boston, USA

Heyi Wang
Illinois Institute of Technology
Chicago, USA

Yumeng Sun
Rochester Institute of Technology
Rochester, USA

Qiyuan Wu
University of California, San Diego
La Jolla, USA

Lian Lian
University of Southern California
Los Angeles, USA

Renhan Zhang*
University of Michigan
Ann Arbor, USA

*Abstract*-This paper addresses the increasingly prominent problem of anomaly detection in distributed systems. It proposes a detection method based on federated contrastive learning. The goal is to overcome the limitations of traditional centralized approaches in terms of data privacy, node heterogeneity, and anomaly pattern recognition. The proposed method combines the distributed collaborative modeling capabilities of federated learning with the feature discrimination enhancement of contrastive learning. It builds embedding representations on local nodes and constructs positive and negative sample pairs to guide the model in learning a more discriminative feature space. Without exposing raw data, the method optimizes a global model through a federated aggregation strategy. Specifically, the method uses an encoder to represent local behavior data in high-dimensional space. This includes system logs, operational metrics, and system calls. The model is trained using both contrastive loss and classification loss to improve its ability to detect fine-grained anomaly patterns. The method is evaluated under multiple typical attack types. It is also tested in a simulated real-time data stream scenario to examine its responsiveness. Experimental results show that the proposed method outperforms existing approaches across multiple performance metrics. It demonstrates strong detection accuracy and adaptability, effectively addressing complex anomalies in distributed environments. Through careful design of key modules and optimization of the training mechanism, the proposed method achieves a balance between privacy preservation and detection performance. It offers a feasible technical path for intelligent security management in distributed systems.

## I. Introduction

With the rapid advancement of digitalization and intelligence, distributed systems have been widely applied in critical fields such as financial services, intelligent manufacturing, cloud computing platforms, and Internet of Things systems[1]. These systems have become the core foundation for building large-scale complex applications due to their high scalability, flexibility, and fault tolerance [2]. However, as the scale and complexity of these systems continue to grow, the security challenges they face have become increasingly severe, especially in the area of anomaly detection. The large number of internal components and the wide distribution of nodes make traditional centralized detection methods inefficient in data collection, processing speed, and privacy protection. These limitations hinder their ability to meet current demands for efficient, real-time, and secure anomaly detection[3].

Timely detection of anomalies is crucial for maintaining the stable operation of distributed systems. Anomalies in such systems are often complex, dynamic, and concealed. They originate from various sources, including hardware failures, software defects, network delays, and malicious attacks. These anomalies frequently exhibit nonlinear and interdependent behavior patterns[4]. Therefore, it is critical to develop detection methods that can efficiently identify anomalies without disrupting normal system operations. Additionally, the physical distribution of system nodes and the involvement of multiple organizations or users raise privacy and compliance concerns. These concerns further limit the applicability of traditional centralized detection methods[5,6].

Federated learning has recently attracted wide attention as a distributed machine learning paradigm. Its key advantage lies in enabling global model training without exposing local data, thereby preserving data privacy and security while supporting cross-node collaborative learning. However, challenges remain in real-world applications. In the context of anomaly detection for distributed systems, the highly heterogeneous data distributions across nodes make it difficult for traditional federated learning methods to capture fine-grained differences in abnormal behavior [7]. Moreover, standard federated averaging approaches often obscure local anomaly features, reducing the model's sensitivity to abnormal patterns[8].

To address these issues, the integration of contrastive learning introduces new potential into federated learning. Contrastive learning enhances a model's ability to distinguish anomalies by learning relationships of similarity and difference between samples. When combined with federated learning, it helps the model focus on potential abnormal patterns while preserving data locality. This approach better adapts to heterogeneous data structures in distributed environments [9]. The combined strategy improves generalization and detection accuracy while mitigating the negative effects of data distribution differences on federated model training. It offers a new solution that balances privacy protection and detection performance[10,11]. In summary, anomaly detection in distributed systems based on federated contrastive learning aims to overcome the limitations of traditional methods in data privacy, model generalization, and anomaly pattern recognition accuracy. This approach seeks to build a more efficient, secure, and scalable detection framework. The research direction aligns with the development trends of distributed intelligent systems and provides both theoretical and technical support for security management in future large-scale collaborative systems. It holds significant practical importance and application value.

## II. METHOD

This study introduces a federated contrastive learning-based method for anomaly detection in distributed systems, designed to balance privacy, accuracy, and robustness. The federated learning paradigm provides a privacy-preserving foundation by enabling distributed nodes to collaboratively train a global model without sharing raw data—a strategy proven effective for time-series anomaly detection in industrial systems [12]. To address the challenge of distinguishing subtle abnormal behavior, the method integrates a contrastive learning mechanism, which enhances the model's ability to discern between normal and anomalous patterns. Each node encodes behavioral data—such as system logs, metrics, and system calls—into embedding representations. Drawing from methods in distributed scheduling and decision making [13], positive and negative sample pairs are constructed locally to promote semantic separation in the feature space. Additionally, building on adaptive resource-aware strategies used in reinforcement learning for distributed systems [14], the method ensures efficient local computation and communication, supporting scalability and responsiveness. These components together form a cohesive architecture that supports precise, real-time detection in complex environments. The detailed model architecture is depicted in Figure 1.

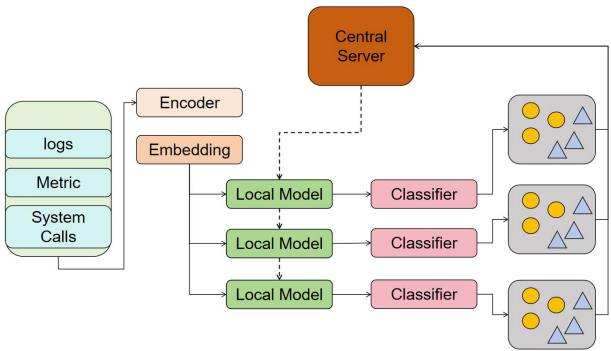

Figure 1. Overall model architecture diagram

In each round of federated training, each local node i uses its local data $D_i = \{x_i^{(1)}, x_i^{(2)}, ..., x_i^{(n)}\}$ to update the model. First, the original behavior data is mapped to the representation space through an encoder network $f_\theta(\cdot)$, that is:

$$Z_i^{(j)} = f_\theta(x_i^{(j)})$$

Following the embedding process, the model constructs positive and negative sample pairs to support contrastive representation learning. Positive pairs consist of samples from the same behavioral category—such as normal-normal or abnormal-abnormal—while negative pairs are drawn from differing categories. This strategy aligns with the pairing schemes used by Wang et al., who employed semantically consistent grouping in reinforcement learning-driven microservice scheduling to enhance learning precision [15]. Training is conducted using the NT-Xent loss, a contrastive loss function designed to minimize the distance between positive pairs while increasing separation from negative ones. This loss formulation draws from techniques in graph neural network-based perception systems, where relational consistency among samples is key to model generalization [16]. To ensure the robustness of these pair constructions across distributed environments, the method incorporates trust-constrained strategies for task coordination, as explored by Ren et al. in their work on policy learning for network traffic scheduling [17]. This ensures that sample pairing remains reliable even under conditions of node heterogeneity and uncertain communication. The contrastive loss function is formally defined as follows:

$$L_{contrast} = -\log \frac{\exp(sim(z, z^+)/\tau)}{\sum_{z' \in Z} \exp(sim(z, z')/\tau)}$$

Where $sim(z, z^+)$ represents the cosine similarity between samples, $\tau$ is the temperature parameter, and $Z$ contains the embedding representation of all positive and negative sample pairs.

To integrate the learning results of multiple local nodes, after each round of training, the local model parameters $\theta_i$ will be uploaded to the central server for aggregation, and the weighted average strategy will be used to update the global model parameters:

$$\theta_{global} = \sum_{i=1}^{N} \frac{|D_i|}{\sum_{j=1}^{N} |D_j|} \theta_i$$

To enhance model adaptability in the face of data heterogeneity across distributed nodes, this method introduces a local regularization term that preserves the sensitivity of each personalized model to distinctive local anomaly features. This term is designed to preserve sensitivity to node-specific abnormal behaviors, enhancing the model's personalization capacity. The need for such localized sensitivity is underscored in prior work on time series transformation for interpretability, where preserving the uniqueness of event-level data patterns was critical to effective mining [18]. The complete loss function incorporates three components: a contrastive loss for embedding discrimination, a classification loss (such as cross-entropy) to guide predictive accuracy, and a local regularization term for maintaining local model fidelity. This formulation draws on principles from Aidi and Gao's temporal-spatial deep learning framework, which emphasizes context-aware forecasting in dynamic cloud environments [19], as well as Cheng's automated feature extraction strategies for multivariate time series modeling using transformers [20]. Together, these components form a comprehensive objective that supports both global consistency and local adaptiveness. The full loss function is defined as follows:

$$L_{total} = L_{contrast} + \lambda_1 L_{class} + \lambda_2 \| \theta_i - \theta_{global} \|^2$$

Where $\lambda_1, \lambda_2$ is the weight hyperparameter of the loss term, which is used to balance the relationship between contrastive learning, classification accuracy, and model consistency.

Through the above method, the system can not only extract representative local behavior features from each node but also capture the commonalities and differences of abnormal behaviors on a global scale, enhancing the adaptability of the detection model in complex distributed environments. The proposed method makes full use of data feature structure and cross-node information while maintaining privacy, providing an efficient, collaborative, and scalable solution for anomaly detection in distributed systems.

## III. EXPERIMENTAL RESULTS

### A. Dataset

This study uses the SWaT (Secure Water Treatment) dataset as the main source of experimental data. The dataset is derived from a real-world industrial control system testbed that simulates the water treatment process of a plant in Singapore. The system consists of multiple physical components, including water pumps, valves, sensors, and PLC controllers. These components collect key indicator data in real-time, such as water level, flow rate, pressure, and chemical properties. The SWaT dataset is one of the benchmark data sources for anomaly detection in distributed industrial systems. It is widely used to evaluate the effectiveness of anomaly detection algorithms.

The dataset contains 11 days of operational data. The first 4 days record normal operations. The last 7 days include various intentionally injected attacks, such as command injection, sensor value tampering, and control logic manipulation. These attacks simulate realistic threats that industrial control systems may encounter. They help assess the model's ability to distinguish anomalies under complex conditions. The data is presented in a time-series format. Each record includes values for multiple physical variables, allowing for multi-dimensional modeling and analysis.

The physical components of the SWaT system operate relatively independently. Data collection points are distributed across different functional zones. This structure makes the dataset well-suited for simulating local nodes in a distributed environment. It provides a natural foundation for building a detection framework that integrates federated learning and contrastive learning. The dataset supports a comprehensive evaluation of the model's adaptability and robustness under heterogeneous nodes, privacy constraints, and diverse anomaly scenarios.

### B. Experimental Results

This paper first conducts a comparative experiment, and the experimental results are shown in Table 1.

Table1. Comparative experimental results

| Method | F1-Score | Precision | AUC |
| --- | --- | --- | --- |
| FedCAC[21] | 86.4 | 84.9 | 91.2 |
| MOON[22] | 88.1 | 86.7 | 92.5 |
| FedProto[23] | 87.3 | 85.2 | 90.8 |
| Ours | 91.5 | 90.2 | 94.7 |

As shown in the comparative results in Table 1, the proposed method outperforms existing mainstream approaches in three key metrics: F1-score, Precision, and AUC. This demonstrates the significant advantage of federated contrastive learning in anomaly detection for distributed systems. In particular, the proposed method achieves an F1-score of 91.5%, which is 5.1%, 3.4%, and 4.2% higher than FedCAC, MOON, and FedProto, respectively. This indicates a better balance between recall and precision. Such improvement is especially important for anomaly detection, where anomalies are often sparse and complex. Traditional methods tend to overfit to one metric. By contrast, the proposed method enhances the model's ability to recognize fine-grained differences through contrastive mechanisms.

The notable improvement in Precision, reaching 90.2%, implies a lower false positive rate in anomaly detection. In distributed systems, false alarms increase diagnostic costs, waste system resources, and may even lead to unnecessary human intervention. Therefore, high precision is paramount. Contrastive learning constructs positive and negative sample pairs to guide the model in learning more discriminative representations between normal and abnormal behaviors. This enhances the classification ability of local models while preserving data privacy. Moreover, the federated framework effectively integrates knowledge from distributed nodes, further improving the generalization capability. This is evident in the consistent improvement in both Precision and AUC.

In terms of AUC, the proposed method achieves 94.7%, which is significantly higher than FedProto (90.8%) and FedCAC (91.2%). This suggests that the method maintains strong classification ability across various thresholds. AUC is a key metric reflecting the overall performance of a model. It directly measures the global ability to distinguish between normal and abnormal samples. While preserving federated privacy constraints, the proposed method strengthens structural representation in the embedding space by incorporating contrastive learning. This enables the model to better handle data heterogeneity across different nodes. Overall, the proposed federated contrastive learning method not only shows clear performance advantages over recent mainstream models but also aligns well with the core requirements of distributed anomaly detection. It addresses multiple challenges, including privacy protection, model discriminative power, and data distribution variability across nodes. The method demonstrates strong adaptability and scalability in practical applications. It provides an effective technical path for building intelligent and secure detection frameworks for large-scale distributed systems in the future.

This paper also gives an analysis of the detection accuracy of the model under different types of attacks, and the experimental results are shown in Figure 2.

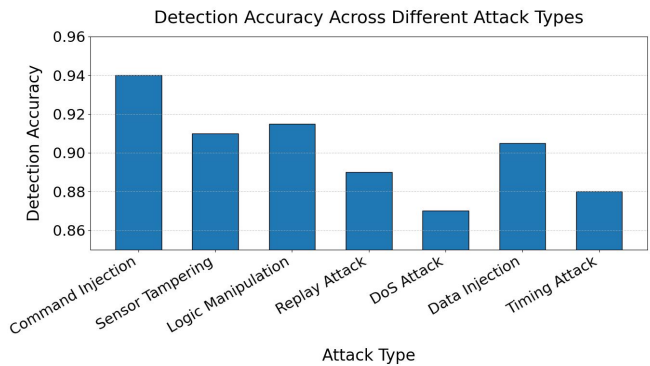

Figure 2. Analysis of the detection accuracy of the model under different types of attacks

As shown in Figure 2, the proposed federated contrastive learning model demonstrates strong detection capability across various attack types. It achieves the highest accuracy under command injection attacks, reaching nearly 0.94. This indicates that the model effectively identifies attack types characterized by clear behavioral changes. Such attacks often result in significant alterations to control flows or device states. These changes produce distinguishable patterns in the embedding space. The contrastive learning mechanism enhances these differences, enabling the model to focus rapidly on abnormal features.

In comparison, the detection accuracy for replay attacks and data injection attacks is slightly lower but still falls within an acceptable range, between 0.88 and 0.90. These types of attacks may not cause immediate or noticeable changes in system behavior. Consequently, they pose a greater challenge for detection. However, by improving sample-level separability in the embedding space, contrastive learning still enables effective identification. Although these attacks introduce relatively weak interference, the model can still detect potential abnormal patterns through abstract representations learned from multiple nodes. This underscores the collaborative advantages of the federated architecture.

It is worth noting that the model performs relatively poorly in detecting denial-of-service (DoS) and timing attacks, with detection accuracy falling below 0.88. These attacks tend to influence data distributions in more indirect ways, often relying on system-wide delays rather than direct manipulation. In distributed environments, where nodes may operate asynchronously, such time-dependent anomalies are more difficult to model consistently. Future work may consider integrating temporal contrastive learning or cross-node contextual fusion mechanisms. These enhancements could improve the model's sensitivity to stealthy timing-based attacks. Overall, the experimental results confirm the wide applicability and strength of the proposed method in identifying multiple types of attacks.

This paper also presents a test of the model's responsiveness in a simulated real-time data stream scenario, and the experimental results are shown in Figure 3.

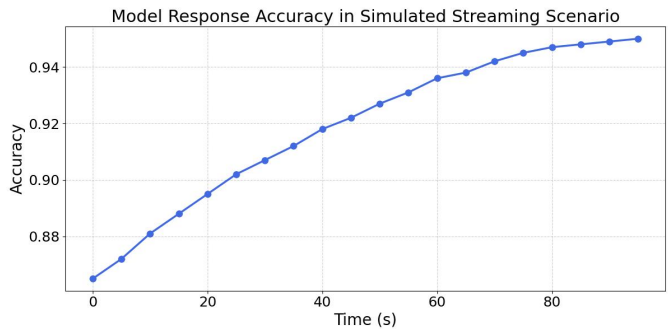

Figure 3. Testing the model's responsiveness in a simulated real-time data stream scenario

As illustrated in the results presented in Figure 3, the proposed federated contrastive learning model exhibits substantial responsiveness within a simulated real-time data stream environment. The model's classification accuracy demonstrates a steady and continuous improvement over time, increasing from approximately 0.87 to nearly 0.95. This consistent upward trajectory evidences the model's ability to incrementally assimilate and refine knowledge of latent abnormal behavior patterns as new data arrives, thereby reflecting its robust online learning capacity and discriminative adaptability. This performance trend further underscores the model's effectiveness in capturing temporal dependencies intrinsic to time series data and in dynamically refining its internal representations in response to evolving patterns. In distributed systems characterized by rapidly fluctuating operational states, static modeling approaches are often encumbered by latency in anomaly recognition and elevated rates of false alarms. In contrast, the integration of a contrastive learning mechanism within the proposed framework facilitates continuous enhancement of feature separability within the embedding space during the data stream, thereby enabling the model to promptly identify and prioritize emerging anomalous behaviors with high fidelity.

Moreover, the experimental findings highlight the practical efficacy and privacy-preserving advantages of the federated learning paradigm in streaming data contexts. By eschewing the need for raw data transmission, individual client nodes are capable of contributing to the global learning process through localized embedding updates in real-time. This architecture enables dynamic, decentralized collaboration while maintaining data confidentiality. The iterative process of local model refinement and federated parameter aggregation results in continuous performance enhancement, thereby substantiating the framework's scalability and robustness under operational conditions reflective of real-world deployments.

IV. CONCLUSION

This paper proposes a novel detection method that combines federated learning and contrastive learning to address the problem of anomaly detection in distributed systems. Traditional centralized detection approaches face increasing challenges related to privacy leakage, data silos, and heterogeneity. In contrast, the proposed method prioritizes privacy while optimizing collaborative modeling and feature discrimination across multiple nodes. By constructing contrastive structures locally and using federated aggregation for global training, the model extracts fine-grained differences in abnormal behavior without exposing raw data. This significantly improves detection accuracy and robustness.

The results show that the proposed method performs well across multiple evaluation metrics and attack types. It accurately identifies common anomaly patterns and is also capable of detecting complex and stealthy attacks. Its performance in a simulated real-time data stream further confirms its responsiveness and adaptability to temporal data. These qualities demonstrate the method's application potential in distributed online monitoring systems. It provides valuable support for security protection in heterogeneous environments such as industrial control systems, cloud platforms, and edge IoT devices. The method offers a practical technical path for real-world system deployment. From a broader perspective, this work advances the integration of federated learning and contrastive learning in the security domain. The proposed framework presents an innovative solution for distributed anomaly detection. It also provides a theoretical foundation for building intelligent detection systems that comply with privacy regulations. The method has potential applications in critical industries, including smart manufacturing, traffic control, power scheduling, and financial risk management. It contributes to improving the overall security and stability of these systems. Additionally, it offers insights for future research on collaboration among low-resource nodes, personalized model adaptation, and self-evolving mechanisms. Future work may focus on analyzing the role of temporal structure in expressing anomaly patterns. Further enhancement may involve integrating graph neural networks, multimodal feature processing, or cross-node relational modeling to improve performance in dynamic and complex environments. To address challenges such as non-independent and identically distributed data and limited communication efficiency, more robust and efficient federated optimization strategies and contrastive designs can be explored. The ultimate goal is to develop a deployable and long-term adaptive intelligent detection framework that can be widely applied across distributed system scenarios.

REFERENCES

[1] A. D. Pazho, G. A. Noghre, A. A. Purkayastha, et al., "A survey of graph-based deep learning for anomaly detection in distributed systems," IEEE Trans. Knowl. Data Eng., vol. 36, no. 1, pp. 1-20, 2023.
[2] Y. Deng, "A Reinforcement Learning Approach to Traffic Scheduling in Complex Data Center Topologies," Journal of Computer Technology and Software, vol. 4, no. 3, 2025.
[3] J. Jithish, B. Alangot, N. Mahalingam, et al., "Distributed anomaly detection in smart grids: a federated learning-based approach," IEEE Access, vol. 11, pp. 7157-7179, 2023.
[4] N. Moustafa, M. Keshk, K. K. R. Choo, et al., "DAD: A Distributed Anomaly Detection system using ensemble one-class statistical learning in edge networks," Future Gener. Comput. Syst., vol. 118, pp. 240-251, 2021.
[5] I. Martins, J. S. Resende, P. R. Sousa, et al., "Host-based IDS: A review and open issues of an anomaly detection system in IoT," Future Gener. Comput. Syst., vol. 133, pp. 95-113, 2022.
[6] Z. Chen, J. Liu, W. Gu, et al., "Experience report: Deep learning-based system log analysis for anomaly detection," arXiv preprint arXiv:2107.05908, 2021.
[7] H. Xin and R. Pan, "Self-Attention-Based Modeling of Multi-Source Metrics for Performance Trend Prediction in Cloud Systems," Journal of Computer Technology and Software, vol. 4, no. 4, 2025.
[8] A. Diro, N. Chilamkurti, V. D. Nguyen, et al., "A comprehensive study of anomaly detection schemes in IoT networks using machine learning algorithms," Sensors, vol. 21, no. 24, p. 8320, 2021.
[9] J. Wei, Y. Liu, X. Huang, X. Zhang, W. Liu and X. Yan, "Self-Supervised Graph Neural Networks for Enhanced Feature Extraction in Heterogeneous Information Networks", 2024 5th International Conference on Machine Learning and Computer Application (ICMLCA), pp. 272-276, 2024.
[10] M. Jain and G. Kaur, "Distributed anomaly detection using concept drift detection based hybrid ensemble techniques in streamed network data," Cluster Comput., vol. 24, no. 3, pp. 2099-2114, 2021.
[11] L. Cui, Y. Qu, G. Xie, et al., "Security and privacy-enhanced federated learning for anomaly detection in IoT infrastructures," IEEE Trans. Ind. Informat., vol. 18, no. 5, pp. 3492-3500, 2021.
[12] H. T. Truong, B. P. Ta, Q. A. Le, et al., "Light-weight federated learning-based anomaly detection for time-series data in industrial control systems," Comput. Ind., vol. 140, p. 103692, 2022.
[13] B. Wang, "Topology-aware decision making in distributed scheduling via multi-agent reinforcement learning," Transactions on Computational and Scientific Methods, vol. 5, no. 4, 2025.
[14] Y. Duan, "Continuous Control-Based Load Balancing for Distributed Systems Using TD3 Reinforcement Learning," Journal of Computer Technology and Software, vol. 3, no. 6, 2024.
[15] Y. Wang, T. Tang, Z. Fang, Y. Deng, and Y. Duan, "Intelligent Task Scheduling for Microservices via A3C-Based Reinforcement Learning," arXiv preprint arXiv:2505.00299, 2025.
[16] W. Zhu, Q. Wu, T. Tang, R. Meng, S. Chai, and X. Quan, "Graph Neural Network-Based Collaborative Perception for Adaptive Scheduling in Distributed Systems," arXiv preprint arXiv:2505.16248, 2025.
[17] Y. Ren, M. Wei, H. Xin, T. Yang, and Y. Qi, "Distributed network traffic scheduling via trust-constrained policy learning mechanisms," Transactions on Computational and Scientific Methods, vol. 5, no. 4, 2025.
[18] X. Yan, Y. Jiang, W. Liu, D. Yi and J. Wei, "Transforming Multidimensional Time Series into Interpretable Event Sequences for Advanced Data Mining," 2024 5th International Conference on Intelligent Computing and Human-Computer Interaction (ICHCI), pp. 126-130, 2024.
[19] K. Aidi and D. Gao, "Temporal-Spatial Deep Learning for Memory Usage Forecasting in Cloud Servers", 2025.
[20] Y. Cheng, "Multivariate Time Series Forecasting Through Automated Feature Extraction and Transformer-Based Modeling," Journal of Computer Science and Software Applications, vol. 5, no. 5, 2025.
[21] X. Wu, X. Liu, J. Niu, et al., "Bold but cautious: Unlocking the potential of personalized federated learning through cautiously aggressive collaboration," Proceedings of the IEEE/CVF International Conference on Computer Vision, pp. 19375-19384, 2023.
[22] Q. Li, B. He and D. Song, "Model-contrastive federated learning," Proceedings of the IEEE/CVF Conference on Computer Vision and Pattern Recognition, pp. 10713-10722, 2021.
[23] Y. Tan, G. Long, L. Liu, et al., "Fedproto: Federated prototype learning across heterogeneous clients," Proceedings of the AAAI Conference on Artificial Intelligence, vol. 36, no. 8, pp. 8432-8440, 2022.